\documentclass[conference]{IEEEtran}
\usepackage[utf8]{inputenc}

\usepackage{graphicx}
\usepackage{subcaption}
\usepackage{xcolor}
\usepackage{verbatim}
\usepackage{svg}
\usepackage[hyphens]{url}

\title{
Distributing Deep Learning Hyperparameter Tuning for 3D Medical Image Segmentation
}

\makeatletter
\newcommand{\linebreakand}{%
  \end{@IEEEauthorhalign}
  \hfill\mbox{}\par
  \mbox{}\hfill\begin{@IEEEauthorhalign}
}
\makeatother

\author{
    \IEEEauthorblockN{Josep Ll. Berral$^{\dagger}$, Oriol Aranda, Juan Luis Dominguez, Jordi Torres$^{\dagger}$}
    \IEEEauthorblockA{Barcelona Supercomputing Center, Universitat Politècnica de Catalunya$^{\dagger}$\\
    josep.berral@bsc.es, oriol.aranda@bsc.es, juan.dominguez@bsc.es, jordi.torres@bsc.es}
}

\begin{document}

\maketitle


\begin{abstract}

Most research on novel techniques for 3D Medical Image Segmentation (MIS) is currently done using Deep Learning with GPU accelerators. The principal challenge of such technique is that a single input can easily cope computing resources, and require prohibitive amounts of time to be processed. Distribution of deep learning and scalability over computing devices is an actual need for progressing on such research field.
Conventional distribution of neural networks consist in "data parallelism", where data is scattered over resources (e.g., GPUs) to parallelize the training of the model. However, "experiment parallelism" is also an option, where different training processes (i.e., on a hyper-parameter search) are parallelized across resources. While the first option is much more common on 3D image segmentation, the second provides a pipeline design with less dependence among parallelized processes, allowing overhead reduction and more potential scalability.
In this work we present a design for distributed deep learning training pipelines, focusing on multi-node and multi-GPU environments, where the two different distribution approaches are deployed and benchmarked. We take as proof of concept the \textit{3D U-Net} architecture, using the MSD Brain Tumor Segmentation dataset, a state-of-art problem in medical image segmentation with high computing and space requirements.
Using the BSC MareNostrum supercomputer as benchmarking environment, we use TensorFlow and Ray as neural network training and experiment distribution platforms. We evaluate the experiment speed-up when parallelizing, showing the potential for scaling out on GPUs and nodes. Also comparing the different parallelism techniques, showing how experiment distribution leverages better such resources through scaling, e.g. by a speed-up factor from x12 to x14 using 32 GPUs.
Finally, we provide the implementation of the design open to the community, and the non-trivial steps and methodology for adapting and deploying a MIS case as the here presented.

\end{abstract}



\IEEEpeerreviewmaketitle

\section{Introduction}

In the past decade, Deep Learning methods (DL) have revolutioned the fields of machine learning, computer vision, and data analytics. This has supposed a huge leap forward in academic research, also in industrial development. Medicine data analysis is one of these fields, leveraging state-of-art Neural Networks for 3D Medical Image Segmentation (MIS), to create models with ground-breaking accuracy and efficiency.
There is a large catalog of DL techniques and models focusing on detection, diagnosis and segmentation of 3D data. The principal challenge is that processing 3D medical data on neural networks is computationally expensive and requires high amounts of memory space. Speeding-up approaches attempt to split data in subpatches before feeding them to the deep network, reducing memory usage requirements but losing spatial information required for good accuracy. In cases targeting the highest accuracy, 3D images need to be processed as full images, then distribution and parallelism are a necessity.

Adapting and tuning state-of-art DL models for Medical Image Segmentation is not trivial. A single model training experiment can be costful in resources and time, not to mention that in most cases hyper-parameter search is also required, multiplying the number of training processes to perform. Scaling out computing resources becomes a necessity, by parallelizing the different experiments and modeling processes across the available resources as computing nodes and GPUs. 
Distribution of the different parts of the training pipeline, including data transformation, data deployment and process placement, must be properly engineered and adapted. Distributed deep learning pipelines can be designed towards "data distribution", where data is distributed along different GPUs for training models as a All-Reduce process, or "experiment distribution", where different GPUs deal with different models in parallel. Previous works explored how data distribution can speed up DL training; however, both approaches can be used to speed-up sets of DL training experiments.

In this work we present the design, methodology and evaluation of Distributed Deep Learning approaches, considering multi-node and multi-GPU environments for scaling out resources. As proof of concept and benchmarking use case, we distribute a state-of-art full-3D volume Medical Image Segmentation network, the 3D U-Net for Brain Tumor Segmentation, on the BSC MareNostrum supercomputing GPU environment. Our methodology pays special attention on data transformation from standard MIS formats (i.e., NIfTI~\cite{nifti}, DICOM~\cite{mildenberger2002}), and uses DL and experiment distribution frameworks like TensorFlow and Ray for both data and experiment parallelism. We evaluate the scalability of the different distribution methods, proving the potential for multi-node and multi-GPU scale out, and comparing the speed-up provided by both distribution methods.

Acceleration of DL pipelines on Medical Image Segmentation are an important medical research use case, in which supercomputing centers are getting more involved day by day. Supercomputing research centers have available large pools of resources to research on High-Performance Computing, including novel architectures and platforms, optimization of high-performance demanding applications, and accelerating Articifial Intelligence and DL use cases. A case like the presented Brain Tumor Segmentation, using 3D U-Net networks with full volume input, is a modelic benchmark for such research where the pipeline distribution is not trivial. Full volume input requires heavy memory usage offered by GPUs, and data must be transformed and arranged for fitting in the device without disrupting the pipeline. And it is known that alternative shortcuts for treating this problem, like subpatching the input dataset, do not perform as good as desired due to the loss of spatial information. Furthermore, full-volume input converges faster, reducing training and inference time.

The evaluation of the presented approaches, and the study of DL distribution scalability are performed in the MareNostrum-CTE GPU Supercomputing environment, composed by state-of-art GPUs NVIDIA V100 16GB deployed on a grid of HPC computing nodes. Results on model dice score (DSC) are kept as reference, to ensure that any pipeline or data modification affects the quality of the resulting models. As a result, we provide the times and speed-up for resource scaling out between 1 GPU and 32 GPUs, in 4-GPU computing nodes. We also observe that in experiment parallelism, distributed components have less dependence than data parallelism, introducing less overhead when scaling out, increasing speed-up from $\times12$ to $\times14$ with respect data parallelism on 32 GPUs.
Finally, the implementation of the presented methodology, with all the details and non-trivial adjustments required for adapting the standard 3D U-Net model and the MSD Brain Tumor dataset, is made public and available to the community, as means to help deploying such workloads on scientific clusters, supercomputers and Cloud environments.



Summarizing, the contributions presented in this paper are:
\begin{enumerate}
    \item The design and methodology for distributing Deep Learning MIS towards full-3D volume input, considering data versus experiment distribution with hyper-parameter search, across multi-node multi-GPU environments.
    \item The adaption of the Brain Tumor Segmentation using 3D U-Net network, as benchmark of speed-up on multi-node and multi-GPU environments. Also, the study and comparison of the speed-up offered by the different distribution approaches.
    \item A framework using TensorFlow and Ray for training and inferente on multi-GPU environments, considering the non-trivial adaptions for full volume 3D imaging, as guide and community open-source software.
\end{enumerate}

This paper is organized as follows: Section~\ref{sec:related_work} introduces the state of the art and background methodologies. Section~\ref{sec:methodology} describes the presented methodology. Section~\ref{sec:experiments} shows the experiments and results validating our approach. Finally, Section~\ref{sec:conclusions} provides some discussion and the conclusions.

\section{State of the Art}
\label{sec:related_work}

\subsection{Related Work}

\subsubsection{Deep Learning on Medical Image Segmentation}
Since introduction of the U-Net in \cite{unet}, this type of network and its variants have achieved state-of-the-art results on various 2D and 3D medical image segmentation tasks \cite{vnet}. Especially in 3D image brain tumor segmentation \cite{brats2017}, the 3D U-Net model based on \cite{3dunet} shown to be better for volumetric data, was used. Numerous approaches of this model have been developed using sampled sub-volume patches \cite{brats2017}, because of memory limitations and to focus more precisely on tumor regions. This last approach, despite it leads to good qualitative results, loses spatial information and it has very poor performing time for both training and inference. Our proposed approach is an end to end solution: using the full volume input, spatial information is not lost, leads to good qualitative results but also better convergence time and hence better scalability.

\subsubsection{Distributed Deep Learning}
In previous works~\cite{Campos2017}, we explored how distributed learning can help to speed up training for neural networks. Several work on spatial, model and data parallelism has been done during the recent years \cite{Chen2014}, \cite{jia2018}, \cite{Shallue2019}, including the implementation of these techniques into the most used deep learning frameworks, e.g. Tensorflow. In spatial parallelism, the data is split into subpatches, and they are sent into different devices with some spatial information. Model parallelism consist on splitting the model and each device is responsable for computing its own piece. In data parallelism, a batch of data is split across the devices and each one computes a mini-batch; it's the most used and is demonstrated as the most efficient and preferred approach whereas either the model or a sample of data can be fed into memory. All these approaches try to solve the common problem of memory limitations when using heavy datasets or models, hence, specially in medical images their application has been also studied \cite{Haryanto2017}. Spatial parallelism has been applied for high resolution medical image analysis \cite{Hou2019}. Model parallelism has been proposed for medical image segmentation \cite{zhu2020}. Data parallelism has been also used for COVID-19 diagnosis based on CT scans \cite{He2020}, text and feature extraction based diagnosis using CNN models \cite{Sierra-Sosa2020}, \cite{Usama2018}. Some studies combine some of the abovementioned techniques, called hybrid parallelism, to handle 3D images and models \cite{Oyama2021}, \cite{akintoye2021}. 
In \cite{Guedria2020} a scalable toolkit for medical image segmentation is presented, but is privative and only two models are provided.
In our method we use a data parallelism approach, and we integrate the pipeline for preprocessing and reading the data.

\subsubsection{Distributed Hyperparameter tuning}
Since the begining when the use of neural networks was first introduced, the hyperparameter optimization or tuning has been essential to improve their performance \cite{Bergstra2011}. It is also well known, that it is a tedious and slow process, for that reason several studies on distributing it and thus increase its performance have been carried out \cite{Ranjit2019}, \cite{Guo2020}. Furthermore, it has also been proposed on medical image diagnosis \cite{Shankar2020}, \cite{Borgli2019}, \cite{Parvathy2021}, but in these studies their focus is not on efficiency.
In our work we propose an easy to use distributed hyperparameter tuning, which leads to a dramatically improvement on performance and more simple usability.

\subsection{Background}

\subsubsection{The 3D U-Net Model}
The 3D U-Net \cite{3dunet} is the most used model for segmentation tasks in medical imaging. At a high level, the network has an analysis and a synthesys path, the encoder and the decoder respectively with four resolution steps each. In the analysis path, each layer contains two $3\times3\times3$ convolutions each followed by a rectified linear unit (ReLu), and then a $2\times2\times2$ max pooling with strides of two in each dimension. In the synthesis path, each layer consists of a transposed convolution of $2\times2\times2$ by strides of two in each dimension, then a concatenation layer followed by two $3\times3\times3$ convolutions each followed by a ReLu. Shortcut connections (concatenations) from layers of equal resolution in the analysis path provide the essential high-resolution features, i.e. spatial information from early layers, to the synthesis path. Referring to the analysis path, the number of filters used in both convolutions at each resolution step are doubled. In turn, the number of filters for the synthesis path is halved.

\subsubsection{Loss Function for Segmentation}
Aside from the architecture, one of the most important elements of any deep learning method is the choice of the loss function. Due to heavy class imbalance (there are typically not many positive regions) the Dice similarity coefficient\footnote{Dice similarity coefficient is known by several names such as Sørensen-Dice Coefficient, Dice's coefficient, Dice index or F1-score} (DSC), which is a measure of how well two contours overlap, is commonly used. Given $A$ and $B$ as sets of voxels, $A$ being the predicted tumor region and $B$ being the ground truth, the Dice index ranges from 0 (complete mismatch) to 1 (perfect match). The model outputs probabilities that each pixel is a tumor or not, and those ouputs are desired to be backpropagated through.
Therefore, an analogue of the Dice index which takes real valued input is utilised \cite{Petronella2005, Herng-Hua2009}. Additionally, the loss function is minimized during training, so it is defined to decrease as performance increases:

\[\mathcal{L}_{Dice}(\hat{y}, y) = 1 - \frac{2 \times \sum_{i, j} \hat{y}_{ij} y_{ij} + \epsilon}{\sum_{i,j} \hat{y}_{ij} + \sum_{i, j} y_{ij} + \epsilon}\]
Where $\hat{y}$ is the prediction mask, $y$ the ground truth mask and $\epsilon$ is a small constant, i.e. 0.1, added to avoid division by zero.
Another variant of the loss, called quadratic Soft Dice Loss following \cite{vnet}, is tested along with the dice coefficient as the metric but seems to lead to worst validation results.

\subsubsection{TensorFlow and Ray API}

TensorFlow and Ray provide APIs to define the training and validation pipelines, towards optimizing the data encoding and distribution, also experiment definition for its distribution.

In Tensorflow, \textit{tf.Data}~\cite{tfdata} provides optimization for pipelines, transparent to the users, allowing also to preprocess data in parallel. E.g., the transformation of raw data into TFRecords, the optimized internal format for TensorFlow data, is directly parallelized through the ``interleave'' and ``map-reduce'' functions. Also, \textit{tf.MirroredStrategy}~\cite{tfdistributed} provides data parallelism across devices on the same machine, creating replicas or copies of a model to be run on different slices of the input data. Note that when using data parallelism, the batch size is divided across devices. Then, to take full advantatge of the available devices, the batch size (and the initial learning rate) must be multiplied by the number of devices used.

For Ray, data parallelism is achieved through \textit{Ray.Cluster}~\cite{raycluster} and \textit{Ray.SGD}~\cite{raysgd} libraries, but over multiple machines instead of devices. Ray handles all the comunication between nodes. In addition, \textit{Ray.Tune} is in charge of performing distributed hyperparameter tuning over the most popular machine learning frameworks. With this, the researcher only needs to focus on the training settings, letting Ray to handle experiment distribution.

\section{Methodology}
\label{sec:methodology}

\begin{figure}[h!]
    \centering
    \includegraphics[width=1\linewidth]{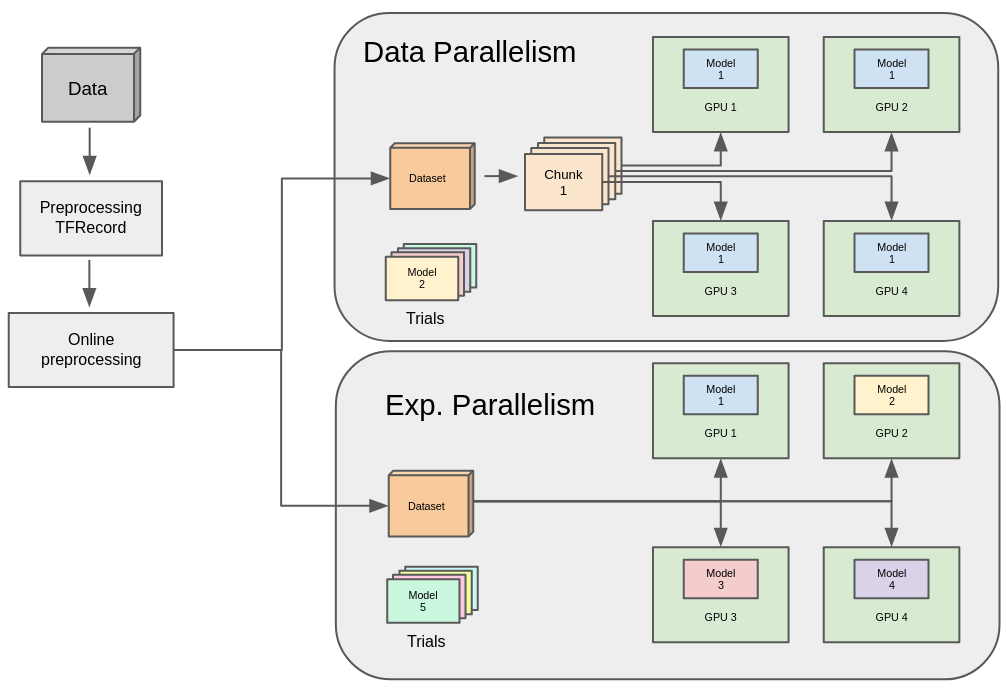}
    \caption{Schema of the different approaches on the proposed methodology}
    \label{fig_pipeline}
\end{figure}

The proposed methodology focuses on two main scenarios: parallelizing data for training, and parallelizing hyper-parameter tuning, both distributing the workload in the two different ways aforementioned. Figure~\ref{fig_pipeline} presents the schema of the created pipeline for the two different approaches.

First approach is the distribution of the training process across multiple computing devices (here GPUs, across a multi-node HPC cluster), leveraging the \textit{Distributed TensorFlow} API to split the experiment data batches across the available resources in each node, then \textit{Ray.SGD} for distributing across nodes. The second approach focuses on the parallelization of the hyper-parameter tuning, using \textit{Ray.Tune}~\cite{raytune} to efficiently distribute the different experiments on hyper-parameter combinations. Both approaches use the training and validation data-sets for training and evaluation respectively on each model.


\subsection{Model Specification}

\begin{figure}[h!tbp]
    \centering
    \includegraphics[width=1\linewidth]{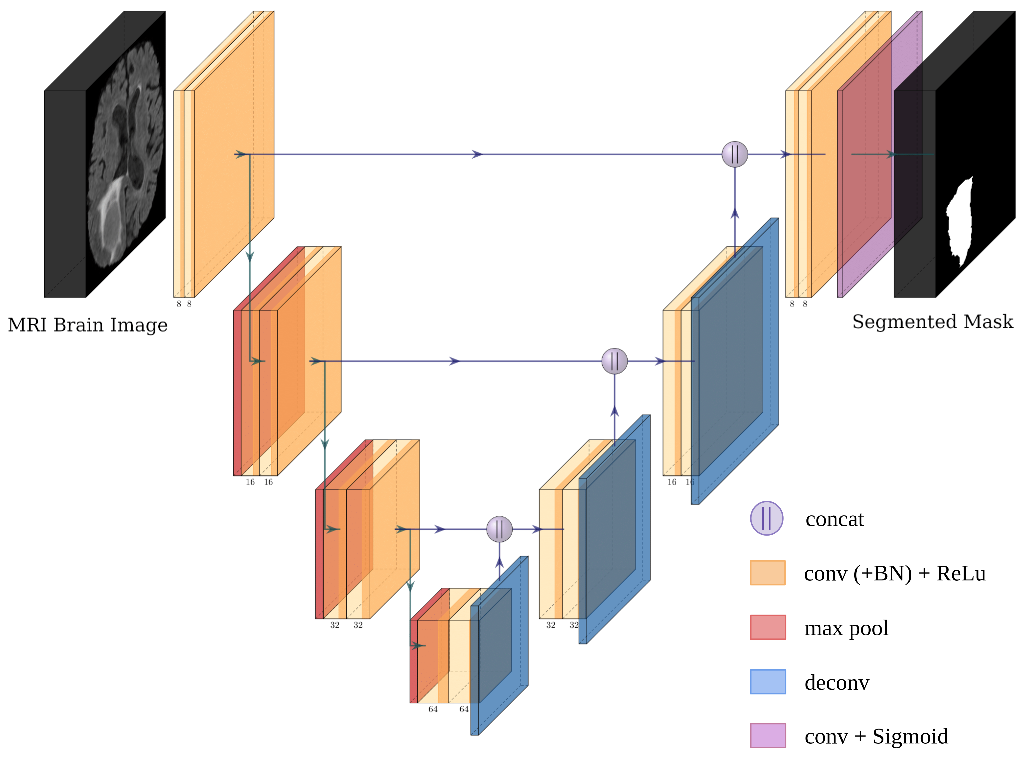}
    \caption{Overview of the used 3D U-Net architecture.}
    \label{fig_3dunet}
\end{figure}

The model used for benchmarking the presented set-up is the previously intrduced 3D U-Net model. As shown in Figure~\ref{fig_3dunet}, the number of filters for each resolution step $s\in\{1, 2, 3, 4\}$ is $8\times 2^{s-1}$. Further, a $1\times 1\times 1$ convolution followed by a sigmoid reduces the number of output channels in the last layer, to match the number of output labels in our case (i.e., 1). Finally, we used \textit{batch normalization} before each ReLU, and a truncated normal kernel initializer for each convolution layer.

As data format, the neural network is built with \textit{Channels First}, being the input a $4\times240\times240\times152$ voxel tile of the image, and the output a $1\times240\times240\times152$ voxel tile matching the ground truth. Once compiled, our benchmarking neural network has 406.793 parameters in total.



\subsection{Architecture Details}

\subsubsection{Adapting the Pipeline}

Either using Ray.Tune for distributing hyper-parameter search or not using it, the used Neural Network engine is TensorFlow (TF). TF uses binarized records as input, i.e. the \textit{TFRecord} format~\cite{tfrecord}, converting images to binary data during the training phase.
An initial analysis on test runs, using the Tensorboard profiler tool~\cite{tfprofiler,tensorboard}, showed us that data loading and its transformation into binary records are the principal bottlenecks in the preprocessing stage of the pipeline, something totally expected from the size and complexity of the target data.
Knowing that the input data will remain the same after each epoch, such data can be binarized off-line before starting the training process. This way, we can avoid to pre-process the data at each epoch, reducing significantly the training cost.
Reading the files for binarization can be parallelized using interleave functions, while the binarization process can be maped over the read data. In addition, the dataset can be pre-fetched.



\subsubsection{Parallelism Levels}

On both approaches, data and experiment parallelism, a hyper-parameter space must be defined with any desired configuration to be performed. In our scenario, this set of configurations becomes the cross-product of the different values for each option in the configuration.

Then, for the data parallelism approach, we can identify three cases when training, depending on the number of available GPUs in the system as $n$:
\begin{itemize}
    \item $n = 1$: Since we only have 1 GPU, training becomes sequential, and each experiment is produced sequentially without any parallelism.
    \item $1 < n \leq M$: Consider $M$ as the number of GPUs on a single computing node ($M = 4$ in our scenario). Here, the Distributed Tensorflow API is triggered to parallelize up to $M$ experiments in parallel in the single computing node.
    \item $n > M$: As more than a single computing node is used, Ray.Cluster is launched to "create" a cluster of available and reacheable resources across physical nodes. With Ray.SGD we apply data parallelism across multiple machines, also Ray handles all the intercomunications and model training updates between nodes.
\end{itemize}


Finally, for the experiment parallelism approach, Ray.Cluster is directly launched to "create" a parallelism cluster along the available resources, then Ray.Tune performs the distributed hyper-parameter tuning process.
Adapting any neural network to Ray.Tune implies adapting its implementation to the standard Ray API, or set of functions for fit, evaluate and predict that Ray expects to find and execute. The basic requirements are to have the training process in a ``training'' function to be called from Ray, having a dictionary containing the hyperparameters as argument. Also, a reporting callback function is required, to provide Ray with the finalization results. Then, the batch of experiments are run through Tune.Run, passing the set of hyper-parameters to explore.




\section{Experiments}
\label{sec:experiments}

\subsection{Dataset}

\begin{figure*}[h!tbp]
    \centering
    \includegraphics[width=\textwidth]{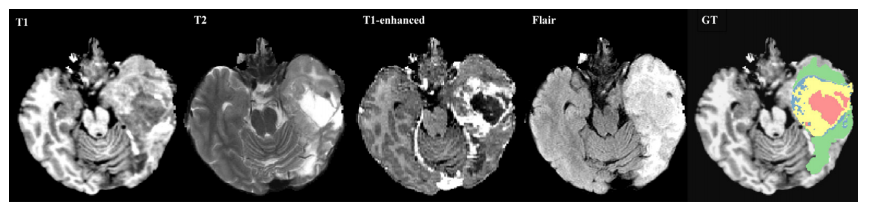}
    \caption{Overview of one sample of the data. The first 4 images corresponds, from left to right, T1w (T1-weighted), T2w (T2-weighted), T1gd (T1-weighted with gadolinium contrast enhancement) and FLAIR (Fluid Attenued Inversion Recovery). The last corresponds to the ground truth.}
    \label{fig_dataset}
\end{figure*}

The dataset used as benchmarking for these sets of experiments is the "Task 1" dataset (brain tumor MRI segmentation), from the MSD challenge~\cite{dataset}. Such dataset consists on 484 multi-modal multi-site MRI data (FLAIR, T1w, T1gd T2w) with 4-class ground truth labels referring to "background", "enhancing" and "non-enhancing tumor", and "edema" segmentations. The volume size for each image is [240, 240, 155], and its resolution/spacing is uniformly $1.0 \times 1.0 \times 1.0 \; mm^3$. Also, for MRI images, the voxel intensities are pre-processed through standardization. Figure~\ref{fig_dataset} shows a sample of the original dataset before pre-processing towards benchmarking, i.e. the four channels and the ground truth image.

The problem corresponding to the original dataset corresponds to a 4-class classification, but for our benchmarking we are reducing the problem to a binary class segmentation task (whole tumer vs. background). The three non-background classes are joined in a single label for "positive", while the background label is considered as "negative". Because of the model architecture, the input dataset must be cropped to sizes of [240, 204, 152] and transposed to "channel first" (from 4-channel inputs) data format. Finally, the dataset is split for training, validation and evaluation as 70\%, 15\% and 15\% respectively.

\subsection{Deployment and Implementation}


The 3D U-Net model is implemented in TensorFlow 2.3, and Ray 1.4.1 is used for the hyper-parameter tuning. The benchmarking pipeline has been deployed in the Barcelona Supercomputing Center MareNostrum-CTE cluster, composed by 52 IBM Power9 (8335-GTH $@$2.4GHz CPUx20) nodes with 4 NVIDIA V100 16GB GPUs each. Infiniband is used as interconnection network. Scalability has been tested for 1 to 32 GPUs (1 to 8 machines).
These are the modules which build the stack of software: gcc/8.3.0 cuda/10.2 cudnn/7.6.4 nccl/2.4.8 python/3.7.4.

One of the principal challenges on these deployments is the volume of each input, against the capacity of the available computing devices. State-of-art GPUs, although having enough resources for common problems, still have very limited memory for use cases like the 3D U-Net. The model is trained with a batch size of 2 inputs per replica, meaning a total batch of $2 \times \#$GPUs when an experiment is distributed.
The number of epochs per experiment is 250, although the model converges much faster and both training and validation are stabilized around epoch 90. The optimizer algorithm used is Adam~\cite{Kingma2017}, with an initial learning rate of $10^{-4}\times$\#GPUs.
Notice here that the learning rate depend on the ratio of data distribution, and because of the scattering of data batches across devices, we need to aproximate it, e.g., by using the Cyclic Learning Rates technique~\cite{Leslie2017}.

\subsection{Performance Analysis}

In order to validate our study and comparison benchmarking, we must ensure that the proposed architecture, deployment and modifications do not affect the performance of the models in terms of correctness (i.e., dice score). For the different experiments here performed, the evaluation on the validation and test sets provide a dice score of 0.89, which are the results of the state-of-art 3D U-Net model. Hence, our methodology and architectures are capable of keeping the dice score results while improving the performance notably.


The following comparison between the two distribution architectures, as seen in Table~\ref{results_tune}, shows the scalability and speed-up provided by doubling the resources (i.e. GPUs), from 1 to 32. Every execution has been run three times, providing here the average.

\begin{table}[h!tbp]
\centering
\begin{tabular}{ |c|c|c|c|c| }
 \hline
 ~ & \multicolumn{2}{c|}{Data Parallel Method} & \multicolumn{2}{c|}{Experiment Parallel Method} \\
 \hline
 \# GPUs used & Elapsed time & Speedup & Elapsed time & Speedup\\
 \hline
 1 & 44:18:02   &   1.00    & 44:20:19	&	1.00 \\ 
 2 & 23:09:28	&	1.91    & 22:24:39	&	1.98 \\
 4 & 15:09:35	&	2.92    & 11:32:20	&	3.84 \\
 8 & 7:41:12	&	5.76    & 7:03:17	&	6.28 \\
 12 & 5:59:59	&	7.38    & 5:35:22	&	7.93 \\
 16 & 4:26:50	&   9.96    & 4:11:54	&	10.56 \\
 32 & 3:21:44	&	13.18   & 2:55:06	&	15.19 \\
 \hline
\end{tabular}
\caption{Results on data parallelism method and experiment parallelism method}
\label{results_tune}
\end{table}

We observe that both architectures follow an almost-linear progression on speed-up when doubling the available GPUs. We must have into account that every computing node has 4 GPUs, and using more implies a communication overhead when distributing a single model across nodes. In comparison, when distributing the hyper-parameter search, each execution is independent from the next one. That prevents overheads from data shuffling or intermediate results communication, as every parallel run is self-contained.

\begin{figure*}[t!bp]
\begin{subfigure}{0.5\linewidth}
    \includegraphics[width=1\linewidth]{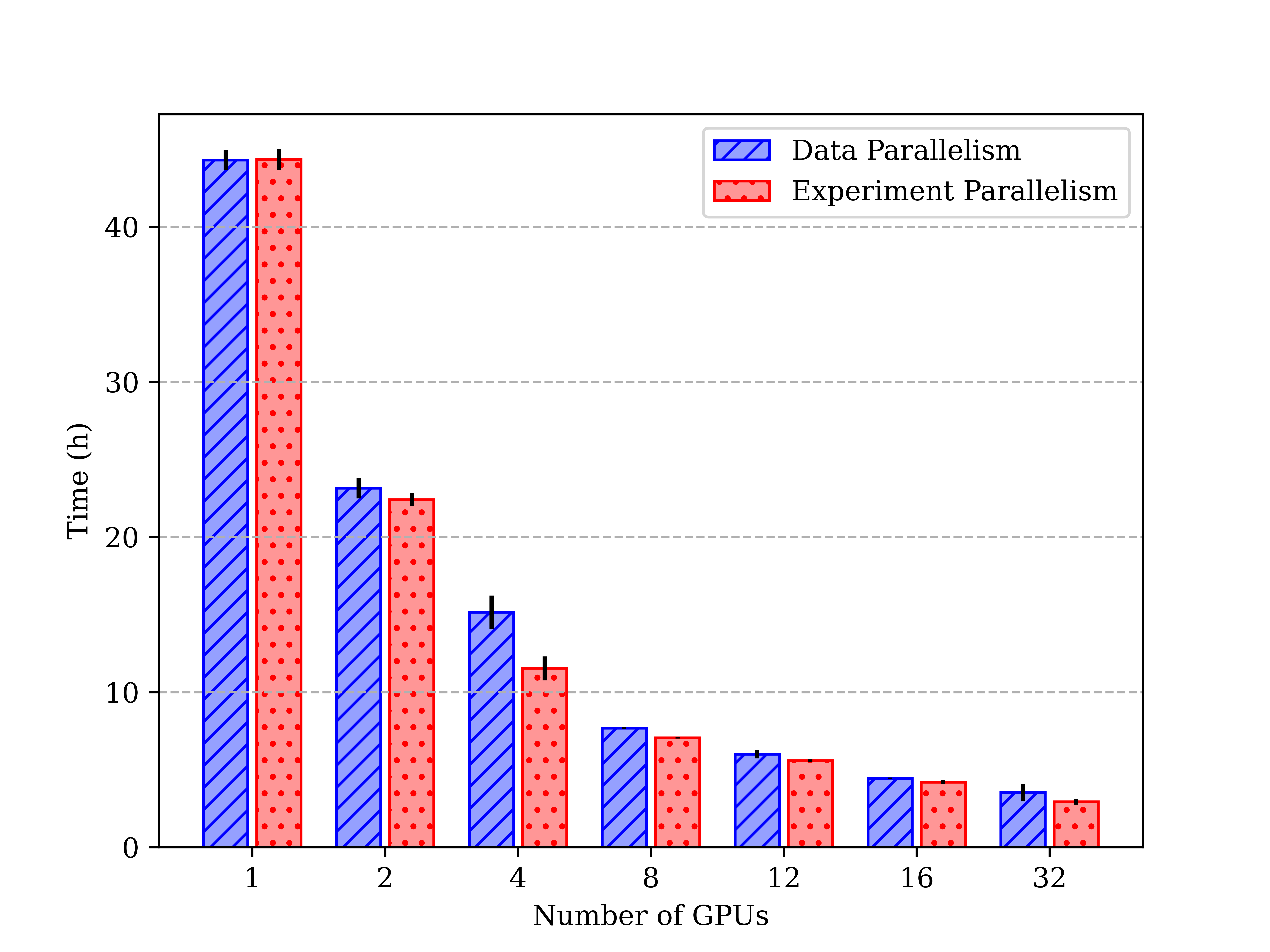}
    \caption{Average elapsed time per number of GPUs, with max and min}
    \label{fig_scalability}
\end{subfigure}
\begin{subfigure}{0.5\linewidth}
    \includegraphics[width=1\linewidth]{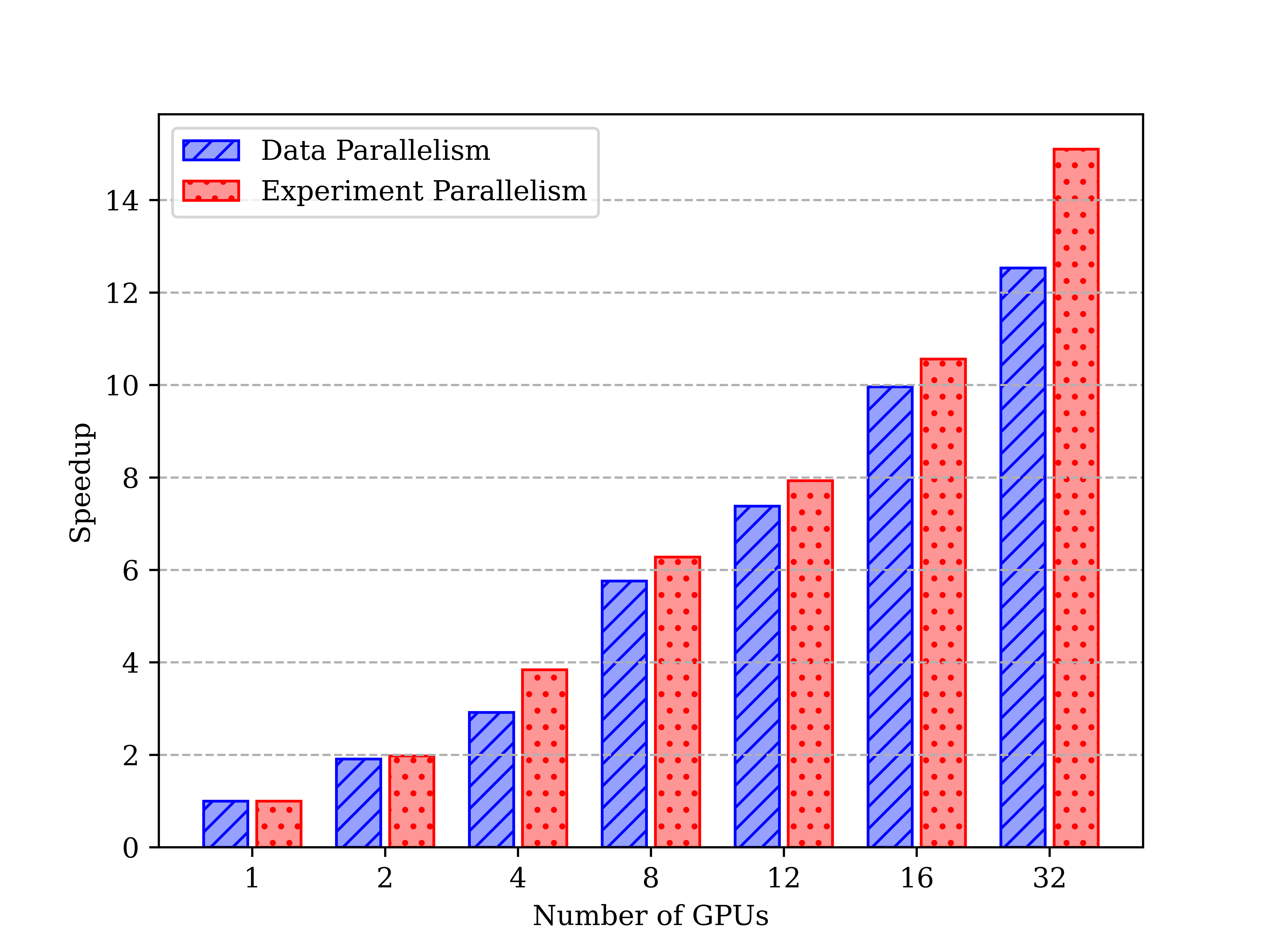}
    \caption{Average speed-up per number of GPUs}
    \label{fig_speedup}
\end{subfigure}
\caption{Comparision of mean elapsed time (a) and speedup (b) between the two methods.}
\label{fig_improvement}
\end{figure*}

Figure~\ref{fig_improvement} displays those results, highlighting the differente between both methods: on the average elapsed time per number of GPUs, and the average speed-up per number of GPUs. Although both methods scale really well, the Ray.Tune method for hyper-parameter distribution shows better improvement on time/speed-up. It is important to recall that, given the amount of experiments that are usually performed when finding the best model for MRI, the smallest improvement on execution time for these kind of experiments can easily add up to hours and days when repeating runs or expanding the hyper-parameter search space.

%

\section{Conclusions}
\label{sec:conclusions}

\subsection{Summary}

In this work we are proposing, studying and evaluating the distribution and scalability of heavy-data workloads, like Neural Network training for Medical Image Segmentation, on multi-GPU and multi-node architectures. This kind of data and problems pose a problem when being deployed on fast-computing devices like GPUs, due to their resource demant and memory occupation of each single input image. Cases like MIS for brain tumor using state-of-art neural networks like 3D U-Net make evident the urgent need for scaling for medical research to leverage deep learning.

Deploying a multi-node multi-GPU architecture provides different options for distributing neural network experiments: data distribution, where data is scattered across available devices to process them in parallel for a single model; and experiment distribution, where each device trains different models from a list of potential model configurations (i.e., hyper-parameter search). Here we provide the system design for both kinds of distribution, with the practical details when deploying a MIS 3D U-Net network as a proof of concept. We highlight the procedures and steps to adapt the problem towards a multi-node multi-GPU cluster (e.g., the BSC MareNostrum supercomputer), and we provide the implementation as open source software as reference for researchers and engineers that require deploying such pipelines.

After benchmarking the proposed architectures using the MSD Brain Tumor Segmentation dataset, with the 3D U-Net image segmentation network, we show the potential of scaling multi-device and node. Also, we compare the two different experiment distribution approaches, in a hyper-parameter search scenario, by serializing experiments while doing data distribution, and by distributing single-device experiments as experiment distribution. This has been done using TensorFlow and Ray platforms for Neural Network training and experiment parallelism.

\subsection{Open Community Framework}
\label{sec:framework}

As initially mentioned, one of the main contributions is to provide the results of this work as an open framework and guide for the community\footnote{Available Open-Source code: \url{https://github.com/HiEST/DistMIS} (Oct'21)}. The principal efforts of this work have been set in finding the correct configurations and capabilities of large-scale computing systems, to show direct improvement of such Deep Learning distribution approach, in a way that these and other use cases can leverage.
The idea behind this is that those users who require Deep Learning in any medical imaging task, can proceed efficiently by following the deployment guide on their systems, and adapt the framework for their purposes (e.g., changing the model architecture and the dataset reader to their desired ones, etc). This will allow users to use the properly integrated pipeline, with the option to automatically perform DL distribution as hyper-parameter tuning or as single experiment training.

\subsection{Future Work}

An important issue, noticed during the scaling over GPU acceleration devices, is the limitation of memory when processing datasets with large inputs, as happens in 3D MIS learning. On scenarios like that, batch sizes are forcefully reduced to 2 or even 1 input, as there is no room in GPU memory for more.
A solution to such problems is to consider model or pipeline parallelism, where the training pipeline for a single model is split across devices. Such distribution is more difficult than data or experiment distribution, since the neural network must be disaggregated.
Next steps focus on scaling resources using model parallelism, to surpass the problem of large input units. Frameworks allowing to distribute models are becoming state-of-art, and being pushed on by GPU manufacturers (e.g., NVidia and DeepSpeed~\cite{nvidia-deepspeed}) showing the potential of such techniques and devices to accelerate Deep Learning even more.


{\small
\section*{Acknowledgements}
This work has been partially financed by the European Commission (EU-H2020 INCISIVE GA.952179, and CALLISTO GA.101004152). Also the Spanish Ministry of Science (PID2019-107255GB-C22/ AEI / 10.13039/501100011033), and Generalitat de Catalunya through the 2017-SGR-1414 project.
}

\bibliographystyle{ieeetr}
\bibliography{main.bib}

\end{document}